\documentclass{article} 
\usepackage{iclr2021_conference,times}

\usepackage{amsmath,amsfonts,bm}

\def\eqref#1{equation~\ref{#1}}

\def\1{\bm{1}}

\DeclareMathAlphabet{\mathsfit}{\encodingdefault}{\sfdefault}{m}{sl}
\SetMathAlphabet{\mathsfit}{bold}{\encodingdefault}{\sfdefault}{bx}{n}

\usepackage{hyperref}
\usepackage{url}
\usepackage{todonotes}

\title{A call for better unit testing for Invariant Risk Minimisation}

\iclrfinalcopy

\author{Chunyang Xiao$^{1,{\dag}}$, Pranava Madhyastha$^{2,{\star}}$
\vspace{2mm}\\
\normalsize{\textit{$^{1}$Bloomberg}}\\
\normalsize{\textit{$^{2}$Imperial College London}}\\
\small{Email: $^{\dag}$ cxiao35@bloomberg.net, $^{\star}$pranava@imperial.ac.uk}
}
\begin{document}

\maketitle

\begin{abstract}
In this paper we present a controlled study on the linearized IRM framework (IRMv1) introduced in~\citet{arjovsky2020invariant}. We show that IRMv1 (and its variants) framework can be potentially unstable under small changes to the optimal regressor. This can, notably, lead to worse generalisation to new environments, even compared with ERM which converges simply to the global minimum for all training environments mixed up all together. We also highlight the isseus of scaling in the the IRMv1 setup. These observations highlight the importance of rigorous evaluation and importance of unit-testing for measuring progress towards IRM.
\end{abstract}

\section{Introduction}
\label{sec:intro}
Invariant risk minimization (IRM)~\citep{arjovsky2020invariant} is a machine learning framework whose primary goal is to learn invariances across multiple training environments. In contrast to Empirical Risk Minimisation (ERM) \citep{10.5555/2986916.2987018}, IRM does not make any assumptions on training samples to be identically distributed; instead, samples are sampled from multiple environments and hence proposes to find associations that are invariant across these environments. 

ERM approaches usually assume the training set is independently sampled from a single distribution and seek to optimize the performance of the machine learning algorithms for samples from the same distribution. In reality, however, training data constitutes samples from a heterogeneous sets of distributions: such as data sampled from different sources, different time, etc.. Thus an ideal algorithm would have learned correlations that are `\emph{true}' correlations which are consistent across environments and hence reduce the chances of capturing spurious correlations that are only correct under a specific environment. 

The IRM framework attempts to find a data  representation that aims to disregard the spurious correlations and optimising `\emph{true}' correlations that are generally correct across the different environments. The general IRM framework~\citep{arjovsky2020invariant} is a non-linear framework which while powerful can be difficult to realise in many scenarios\footnote{For more details, we refer the reader to the discussion in~\citet{arjovsky2020invariant}}. 

In this paper, we focus on a simpler, linearized variant of IRM~\citep{arjovsky2020invariant} also called IRMv1 which is defined with respect to risks. The \emph{risks} in this context are defined on the basis of population errors that include both training errors and feature stability penalization in contrast to solely training errors in ERM. In this paper we perform a controlled study of the linearized setting, which is one of the prominent approaches explored in the recent work~\citep{ahuja2020linear,koyama2021invariance}, and highlight that IRM is highly unstable over the initial settings of the optimal regressor especially for benchmark tasks. Our observations strongly recommend for a more rigorous unit-testing and evaluation setup to correctly estimate progress towards this framework. Our code can be found at \url{https://github.com/chunyangx/IRM\_research}. 

\section{Method}

\subsection{Background}

ERM is the de facto machine learning framework that is used out of box in most of the current models. Briefly it works by building a model which chooses the hypothesis space that minimizes the risk. However, it has been noted that this standard formulation and procedure is not without issues, for example the model tends to learn biases from the dataset, as shown by both simple but illustrating problems~\citep{beery2018recognition} and real problems~\citep{4425549d31d5432ba2cdd896022be8ad, 10.1109/CVPR.2011.5995347}.

One promising direction that attempts to address such issues is to bias the model to learn causal relations and ignore spurious correlations. IRM \citep{arjovsky2020invariant} is one such framework that attempts to achieve this. The basic intuition behind IRM is that causal features are stable across different learning settings (also called environments) while spurious features tend to have larger variance across environments; thus by retaining and focusing only on the stable features across environments, a model trained in the IRM framework can potentially retain causal components and thereby can generalize better across various testing settings.

We present the simplified IRM objective as proposed in~\citet{arjovsky2020invariant} in Equation~\ref{eqn:IRM} which is composed of two terms.
\begin{equation}
\min_{\Phi: \mathcal{X}\to\mathcal{Y}} \sum_{e\in\mathcal{E}_\text{tr}}  \mathbf{R}^{e}{(\Phi)}   + \lambda\|\nabla_{w{\mid}w=1.0}\mathbf{R}^{e}(w\cdot\Phi)\|^{2};
\label{eqn:IRM}
\end{equation} 
where, $\mathbf{R}^{e}{(\Phi)}$ corresponds to the standard ERM framework and minimizes the risk across all environments. Whereas,  $\|\Delta_{w{\mid}w=1.0}\mathbf{R}^{e}(w\cdot\Phi)\|^{2}$ penalizes features that have large gradient\footnote{We note that the gradient computed based on a fixed uniform classifier, which is in contrast to usual settings where the weights of the classifier is learned via optimization thus is non-fixed.}. Intuitively, a stable feature once optimized will converge to a representation where the gradient tends to zero; non-stable features will have gradients which differ a lot across environments, and will thus be penalized by the second term, making the feature's norm tend to zero and thus ignored by the final model through optimization. 

\subsection{An IRM testing setting}

To test the empirical performance of IRM algorithm, several simple but illustrative settings have been proposed and compared in the literature. In this work, we focus on one graphical regression setting, the graphical model is depicted in Figure~\ref{fig:settings}. The setting is proposed in the IRM original paper \citep{arjovsky2020invariant} and are discussed and bench-marked in  many followup works that include~\citep{ahuja2020linear, aubin2021linear, koyama2021invariance}. 


\label{sec:method}

The setting consists of a simple regression setting with spurious variables. 
As shown by Figure~\ref{fig:settings}, $Z_1^e$ causes $Y^e$ across environments, contrary to $Z_2^e$ which is ``caused" by $Y^e$ and thus is treated as spurious features in the learning setting.\footnote{As we can see in subsection 3.1.2, the main difference between $Z_1^e$ and $Z_2^e$ is not the arrow direction but the noise applied to these variables with the corresponding implications.} This main chain of variables can be further complicated by introducing a co-founder hidden variable $H^e$. Finally, two noise models namely homoskedastic and heteroskedastic with different scales are applied to $Z_1^e$ and $Z_2^e$ respectively.  Thus in total, We consider a total of four different settings.\footnote{We note that this is the same set of settings considered in \cite{ahuja2020linear}, which omits the weight matrices transformation cases compared to \cite{arjovsky2020invariant}.}

Compared to the graphical model as specified in Equation~\ref{eqn:IRM}, we observe that   
previous work on IRM \citep{ahuja2020linear, aubin2021linear, koyama2021invariance} have further simplified the assumptions where the variable $Y^{e}$ to regress to is reduced to a \textit{scalar} which is considered as the sum of all its coordinates. This is usually performed to facilitate model analysis. In this case, it is easy to show that the optimal weight turns into a vector which is essentially the sum of corresponding weight in $W_{1\rightarrow y}$. In this paper, we call this vector the optimal regressor.

\section{Experimental Investigation}
\label{sec:experiments}

\subsection{On experimental settings} 
In this section, we focus on the IRMv1 experimental setting and explore in two directions. 
\begin{figure}[t]
\begin{center}
\includegraphics[scale=0.35]{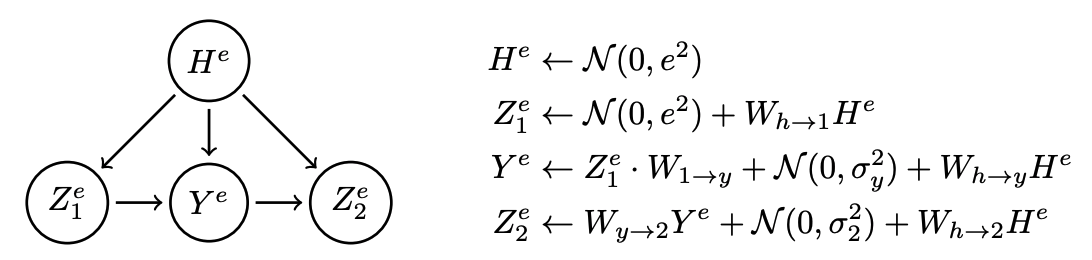}
\caption{The unit test considered in this work.}
\label{fig:settings}
\end{center}
\end{figure}

\paragraph{The problem with scaling}

In the original IRM experimental setup~\citep{arjovsky2020invariant}, the optimal regressor is \textit{set} to a uniform vector of \texttt{ones}. The same setup is further used in the follow up works such as \cite{ahuja2020linear}. While \cite{aubin2021linear} show that IRM is able to generalise sufficiently well under a wider range of settings by sampling the weight matrix from Gaussian distributions, we however notice that the Gaussian distributions in their experimental framework do not vary in scales as we do in this work.  
In this paper we focus on the effects of changes in the results and/or the conclusions based on simply drawing from the Gaussian samples with a small change in scale. 

Our focus on the scales are first inspired by the results that gradient based algorithms have been shown to have difficulties finding optima as they are sensitive to scaling issues \citep{beery2018recognition}. But our empirical results show that IRM formulation seems to particularly suffer from the scaling issues, which can not be simply explained by the optimization aspect. 


In the following we first analyse the experimental setup before present a hypothesis that reflects on this instability.

\paragraph{Noise with multiple roles}

There are two noise terms in the equations shown in Figure~\ref{fig:settings}.\footnote{To simplify the discussion, we will omit the hidden variables $H^e$ in the discussion of here, although one can note that this doesn't lose generality as we focus only on the noise terms.} The noise $\sigma_y^2$ is the natural Gaussian noise that we see in linear regression.
However, we note that the noise $\sigma_2^2$ plays quite a different role as shown in in Figure~\ref{fig:settings}.  To examine this further, we first inverse the arrow such that the equation confirms to the normal prediction based formulation as the arrow now points from target variable $Y$ to spurious features $Z_2$ (as seen in the graph). Let $W_{y\rightarrow2}^{-1}$ the inverse of the matrix $W_{y\rightarrow2}$, by omitting the $H^e$ term, we get:

\[ Y^e \leftarrow W_{y\rightarrow2}^{-1} (Z_2^e - N(0, \sigma_2^2))\]

Thus, the noise term now adds noise in the features instead of targets as $\sigma_y^2$. Similar effects of feature noise has been recently studied in \citep{DBLP:journals/corr/abs-1911-09876} where the authors show that it seems to play very different roles compared to the target variable noise. For simplicity, we omit the derivation here and directly use the result in linear setting, i.e., if the noise term $N(0, \sigma_2^2)$ has a covariance matrix whose norm is significantly larger than $W_{y\rightarrow2}^{-1}$, then the optimal weight will tend to zero (i.e. a vector of zeros). 

Therefore, although in this paper, we propose to test a larger variety of unit test cases for the ground truth model, we observe that only a small subset is sufficient for such testing. For the cases where $W_{y\rightarrow2}^{-1}$ has a small covaraince matrix, the model will naturally converge to weights with smaller magnitudes and thus seem to ignore the spurious features.\footnote{We conducted preliminary experiments where each term in the weight matrix is sampled form $G(0,1)$ and the results do show that the weights will naturally tend to zero.}

\begin{figure}[t]
\begin{center}
\includegraphics[scale=0.4]{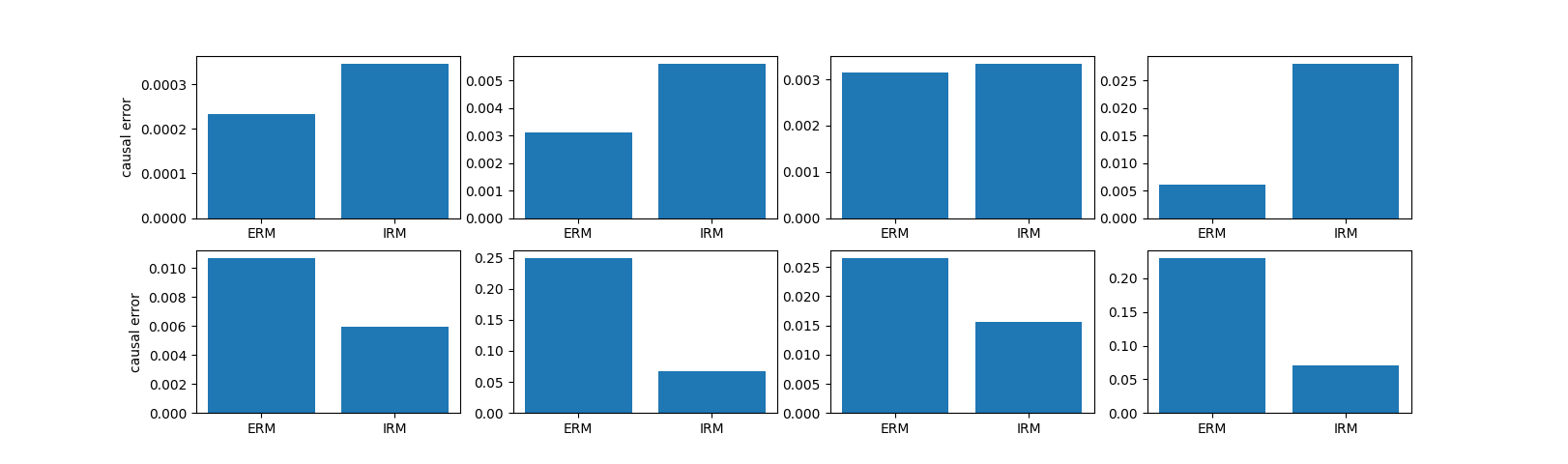}
\caption{Results comparing IRM and ERM on synthetic datasets under various settings. The first shows the results under various settings when the Gaussian is sampled from $G(0,0.1)$ and the second row shows the results under the same settings when the Gaussian is sampled from $G(0,0.35)$.} 
\label{fig:IRMvERM}
\end{center}
\end{figure}

\subsection{Experiments}
Based on these observations, we propose to conduct tests on the similar settings as \cite{ahuja2020linear} but with the only difference where we also consider varying the scale of the ground truth weight matrix $W_{y\rightarrow2}$ and $W_{1\rightarrow y}$ respectively. In our setting, instead of fixing the weights, each term in the matrix is always sampled from a Gaussian distribution; we consider two distributions for the ground truth matrices which are  sampled from $G(0,0.35)$ and $G(0,0.1)$. We believe this is a rather realistic setting as the ground truth scale can differ mildly for different real world applications even under normalized features. 

We choose these two ground truth settings specifically to test the scale issue, on $G(0, 0.35)$ cases, there is a $~70\%$ chance that the each ground truth regressor term falls in the range of $[0.5, 1.5]$. This is similar to the scale in the original setting. In the $G(0, 0.1)$ setting, the sum of weights tend to shrink to a much smaller number and there is almost zero chance that a regressor term lies between $[0.5, 1.5]$.

\subsection{Algorithms under consideration and Results}

In this paper we restrict our empirical investigations to IRM and ERM under the settings described above. In this linear setting, we present the ERM results through both the analytical form\footnote{Similar to \cite{arjovsky2020invariant}, we use an off the shelf \texttt{scikit} learn solver for this.} and SGD optimization. Our empirical observations confirm that under the mild scale change, SGD optimization are still largely capable of convergence under ERM settings, achieving the same performance as the analytical solution.

Our main results are shown in Figure~\ref{fig:IRMvERM}, we observe that under $G(0,0.35)$ settings, IRM outperforms ERM in all the cases. This is consistent to the previous work \citep{arjovsky2020invariant,aubin2021linear}.
However, when we change the samples to have optimal regressor's weight-norms further away from 1, we obtain markedly opposite results. When the weights are sampled from $G(0,0.1$, we observe that for all the considered settings, IRM is outperformed by the simple ERM algorithm.   
\label{ssec:algos}

\section{Discussion}
Our empirical observations show that, when the optimal regressor deviates mildly from the dummy vector set by the IRM (i.e. a vector where each dimension is norm 1), IRM seems to find worse solutions than ERM. However, when the norm is close to the dummy vector, it outperforms ERM. Our observation might be related to other observations\footnote{We point the reader to https://github.com/reiinakano/invariant-risk-minimization for IRM replication experiments.} showing that IRM is difficult to optimize, and fall easily to local minimum when $\lambda$ balancing the feature stability term and ERM term is not well tuned. On the theoretical side, our results might be related to \cite{kamath2021does} which shows that IRMv1 indeed performs worse than ERM depending on the ground truth models.

Our observations suggests that a careful analysis with better unit tests is paramount to better understanding and design of frameworks. A careful setting in simple and well understood environments can benefit real settings \citep{koh2021wilds}. However, we note that in these settings, \cite{koh2021wilds} show that current popular methods including IRM and Group DRO~\citep{DBLP:journals/corr/abs-1911-08731, hu2018does} do not consistently outperform ERM.

\bibliography{iclr2021_conference}
\bibliographystyle{iclr2021_conference}



\end{document}